\documentclass{article}

\usepackage[final]{neurips_2026}
\makeatletter
\renewcommand{\@noticestring}{2nd Workshop on Principles of Generative Modeling (PriGM) at NeurIPS 2026, Paris, France.}
\makeatother

\usepackage[utf8]{inputenc}
\usepackage[T1]{fontenc}
\usepackage{hyperref}
\usepackage{url}
\usepackage{booktabs}
\usepackage{amsmath}
\usepackage[expansion=false]{microtype}
\usepackage{xcolor}
\usepackage{graphicx}
\usepackage{float}
\usepackage{tikz}
\usetikzlibrary{arrows.meta,positioning}
\hypersetup{hidelinks,hypertexnames=false}
\newcommand{\runin}[1]{\par\vspace{2pt}\noindent\textbf{#1}\ }

\definecolor{attentionblue}{HTML}{2C7FB8}
\definecolor{ssmorange}{HTML}{E67E22}
\definecolor{diagramgray}{HTML}{5B6573}
\definecolor{lightblue}{HTML}{EAF3F8}
\definecolor{lightorange}{HTML}{FDF0E5}
\tikzset{
  diagram node/.style={draw=diagramgray!75, rounded corners=1.5pt,
    fill=white, minimum height=0.40cm, inner xsep=2.5pt, inner ysep=1.5pt,
    font=\scriptsize\sffamily},
  question node/.style={diagram node, draw=attentionblue, fill=lightblue},
  distractor node/.style={diagram node, fill=diagramgray!7, minimum width=0.42cm},
  gold node/.style={diagram node, draw=ssmorange, fill=lightorange,
    text=ssmorange!80!black, minimum width=0.42cm},
  diagram arrow/.style={-{Latex[length=1.4mm]}, line width=0.45pt,
    draw=diagramgray!85},
  metric node/.style={diagram node, draw=attentionblue, fill=lightblue}
}

\title{Mixing Matters: Evidence Position Bias Across Sequence Mixers in
Long-Context Question Answering}

\author{%
  Aman Behera\\
  IIT Roorkee
  \And
  Namit Solanki\\
  AISSMS IOIT, Pune
  \And
  Mehul Anand\\
  Independent Researcher
}

\begin{document}
\raggedbottom

\maketitle

\begin{abstract}
Long-context models often use answer-bearing evidence less effectively when it appears away from a prompt boundary.
We compare this position bias across Transformer, state-space, and hybrid Mamba architectures by moving one gold document through a fixed set of nine distractors.
The evaluated Transformers show a beginning advantage that the evaluated state-space models lack, while both families favor evidence near the end.
A matched pure-versus-hybrid Mamba comparison points in the same direction but remains statistically uncertain.
The family difference appears among models that can answer the task.
Transformer primacy reappears on synthetic retrieval, where state-space ceiling prevents a family comparison.
A corpus swap within one state-space architecture changes overall accuracy without a detectable change in curve shape.
Prompt changes can create or nearly remove a beginning advantage.
Position remains decodable across architectures even when response curves differ, and attention concentration near initial tokens covaries with beginning sensitivity in the tested Transformer family.
These exploratory results show that effective context use depends on the evaluated architecture family, capability, task, and prompt, without isolating the sequence mixer as the sole cause.
\end{abstract}

\section{Introduction}

Retrieval can place the right document in a prompt without ensuring that a model will use it.
In multi-document question answering, accuracy often drops when relevant evidence moves from the context edges toward the middle \citep{liu2024lost}.
This gap between available and effectively used context matters for retrieval-augmented systems and motivates robust long-context evaluation \citep{bai2024longbench,hsieh2024ruler}.

Most evidence-position studies focus on Transformers.
Dense attention can mix information directly across token positions, whereas state-space models carry the prefix through recurrent state \citep{gu2023mamba,dao2024mamba2}.
We ask whether these sequence-mixer families exhibit the same position curve.
Across our evaluated systems, they do not.
Transformers favor evidence near the beginning, the tested state-space models lack the same beginning advantage, and both families favor evidence near the end.
The difference emerges only once models can reliably answer the task.

The curve is not an architecture-only signature.
A matched hybrid comparison is directionally consistent but uncertain, a corpus swap changes overall accuracy without detectably changing curve shape, and prompt changes can create or nearly erase the beginning advantage.
Internal measurements further suggest that position information can remain available even when it is not used equally across positions.
Our contribution is therefore an evaluation result: long-context position bias varies across the tested sequence-mixer families and evaluation settings, while current evidence does not establish a causal mechanism.

\section{Paired position evaluation}
\label{sec:method}

\runin{Task}
We use the released Lost-in-the-Middle ten-document question-answering setting, derived from Natural Questions \citep{liu2024lost,kwiatkowski2019nq}.
Each question has one gold document and nine distractors.
For every question, we render ten prompts that hold the question, distractors, and token budget fixed while moving only the gold document.
A no-gold prompt measures the floor and a gold-only prompt measures answerability.
All reported ten-document results draw from an 800-question exploratory allocation or a subset; the held-out confirmatory split remains unopened.

\begin{figure}[H]
\centering
\includegraphics[width=0.78\linewidth]{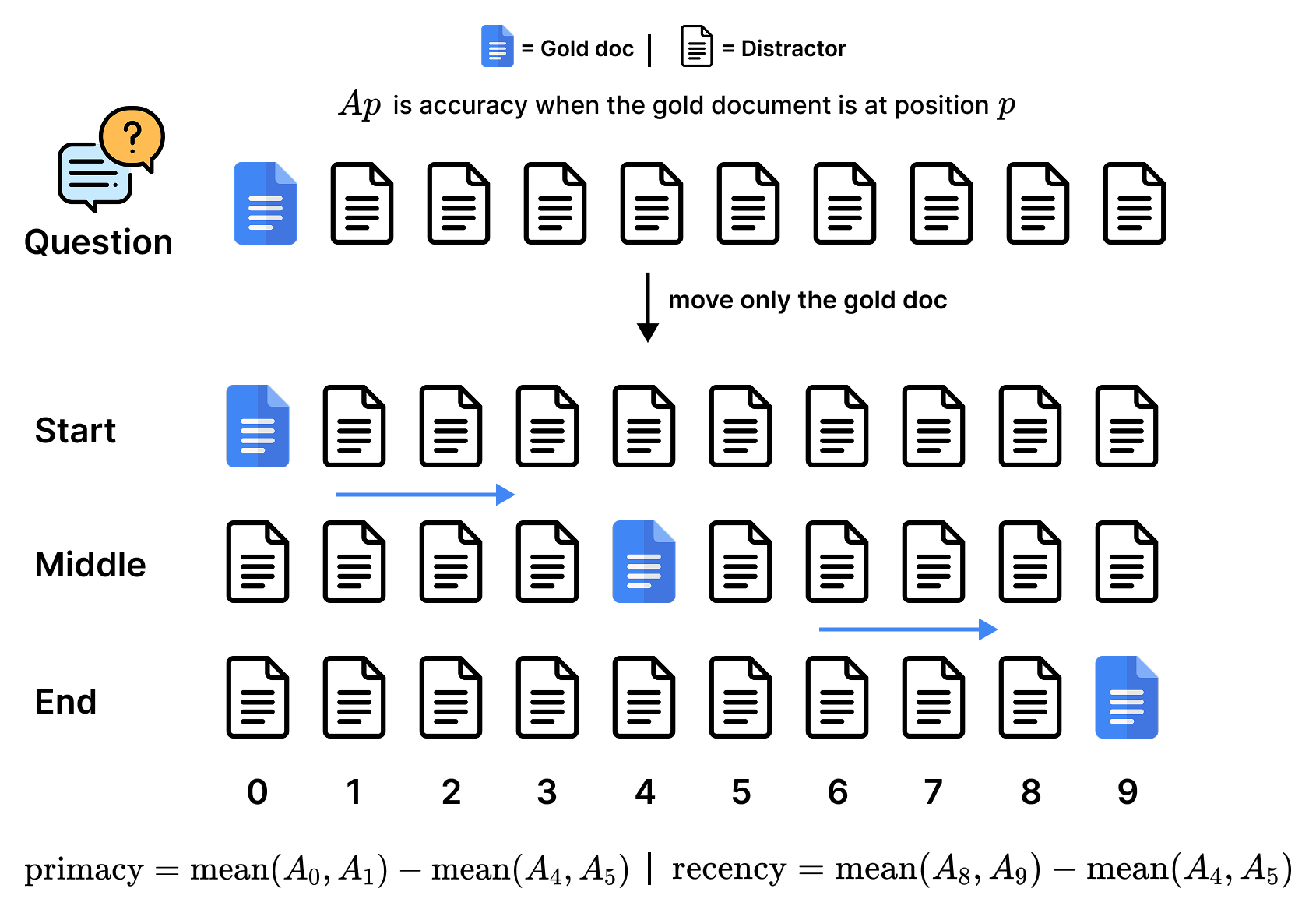}
\caption{Paired position intervention.
$A_p$ is accuracy when the gold document is at position $p$.
Complete ten-position question bundles are resampled together.}
\label{fig:method}
\end{figure}

\runin{Metrics and inference}
Primacy is start-minus-middle accuracy and recency is end-minus-middle accuracy, as defined in Figure~\ref{fig:method}.
The question is the unit of analysis.
We draw 10{,}000 paired bootstrap samples of complete ten-position bundles, report percentile 95\% confidence intervals, and apply Holm correction within each comparison family \citep{efron1979bootstrap,holm1979multiple}.
Intervals quantify variation across questions for fixed prompts and checkpoints, not training-seed variation.
Generation is greedy, and the harness preserves prompts, outputs, failures, token counts, model revisions, and software settings.

\runin{Controls}
On a 200-question exploratory Transformer subset, a released key-value task produces the expected edge effect, sham-gold primacy is near zero, and permuting distractor order changes mean accuracy by at most $0.023$.
These controls check the paired harness in that setting only; they do not cover the matched 8B checkpoints.

\section{Findings}
\label{sec:findings}

\begin{figure}[H]
\centering
\begin{minipage}[t]{0.49\linewidth}
\centering
\includegraphics[width=\linewidth]{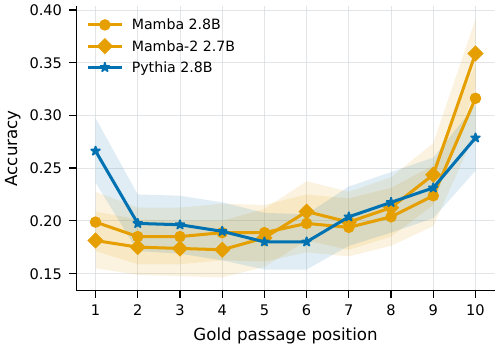}
\textbf{(a) Shared-corpus family comparison}
\end{minipage}\hfill
\begin{minipage}[t]{0.49\linewidth}
\centering
\includegraphics[width=\linewidth]{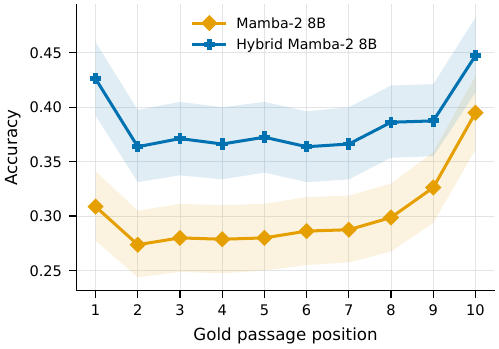}
\textbf{(b) Matched pure and hybrid comparison}
\end{minipage}
\caption{Accuracy as the gold document moves through ten positions on 800 exploratory questions.
Shaded bands are 95\% question-bootstrap confidence intervals.}
\label{fig:curves}
\end{figure}

\runin{Sequence-mixer families differ at usable capability}
In a shared-corpus comparison at similar scale, the Transformer has a positive primacy edge, while two state-space variants do not; all three have positive recency edges (Figure~\ref{fig:curves}a).
Paired interactions confirm that the Transformer-versus-state-space primacy differences are unlikely to arise from question sampling alone.
Across five approximate size pairs, the family gap is absent at the two smallest scales, where answerability is poor, and appears in the three larger pairs.
This is a descriptive family-and-capability trend, not a causal scale threshold.

\runin{The tighter comparison is suggestive, not decisive}
Pure and hybrid state-space checkpoints share training data, tokenizer, scale, depth, and positional setup; the hybrid adds a minority of attention layers \citep{waleffe2024empirical}.
The hybrid shows a larger within-model primacy edge, but the paired hybrid-minus-pure confidence interval includes zero (Figure~\ref{fig:curves}b).
This direction agrees with the family comparison but does not isolate a causal attention effect.

\begin{table}[H]
\centering
\caption{Primary edge effects.
The final row is the paired hybrid-minus-pure interaction; other rows are within-model edges.
For that paired row, Holm $p=0.144$ for primacy and $p=0.089$ for recency, so neither effect survives correction.}
\label{tab:main-results}
\scriptsize
\setlength{\tabcolsep}{3.2pt}
\begin{tabular}{@{}lcccccc@{}}
\toprule
System or contrast & Primacy & 95\% CI & Recency & 95\% CI & Floor & Ceiling \\
\midrule
Transformer & $+0.052$ & $[0.032,0.072]$ & $+0.075$ & $[0.054,0.097]$ & $0.091$ & $0.640$ \\
State-space v1 & $-0.001$ & $[-0.016,0.014]$ & $+0.077$ & $[0.058,0.097]$ & $0.115$ & $0.615$ \\
State-space v2 & $-0.018$ & $[-0.034,-0.003]$ & $+0.105$ & $[0.086,0.124]$ & $0.128$ & $0.619$ \\
Hybrid minus pure & $+0.019$ & $[-0.006,0.044]$ & $-0.028$ & $[-0.055,-0.001]$ & -- & -- \\
\bottomrule
\end{tabular}
\end{table}

\begin{figure}[H]
\centering
\includegraphics[width=0.95\linewidth]{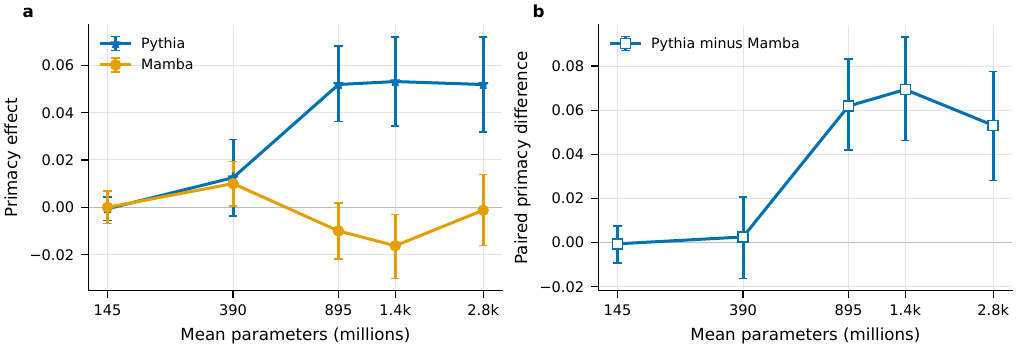}
\caption{Primacy across five approximate size pairs.
The family gap appears only among systems that can answer the task, making this a capability-and-family trend rather than a causal scale result.}
\label{fig:main-scale}
\end{figure}

\section{Discussion and limitations}

\runin{Mechanism measurements motivate hypotheses}
Position is linearly decodable from intermediate states in both families relative to shuffled-label controls, even when their response curves differ.
Within the tested Transformer family, attention concentration near initial tokens covaries with primacy across scale \citep{xiao2024efficient}.
The probe does not show that answer content is stored or used, scale confounds the attention association, and no valid sink-blocking intervention was completed.
The measurements therefore motivate an evidence-utilization hypothesis but do not establish its mechanism.

\begin{figure}[H]
\centering
\includegraphics[width=\linewidth]{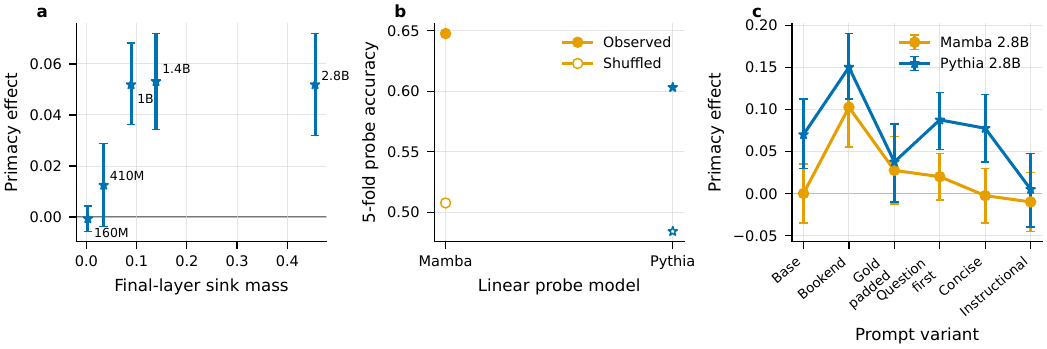}
\caption{Exploratory evidence on attention concentration, position decodability, and prompt sensitivity.
The sink and probe measurements are correlational, and the prompt variants change multiple prompt properties rather than isolating one causal mechanism.}
\label{fig:mechanisms}
\end{figure}

\runin{Implications}
The central finding is that available context length and effective evidence use are different properties.
Models with the same ten documents can exhibit distinct position curves, and prompt layout can alter those curves as much as family choice.
Long-context evaluations should therefore report full position curves, hold distractors fixed, use paired question-level inference, and test prompt sensitivity rather than relying on a single aggregate accuracy.

\runin{Reproducibility}
The released materials record the dataset split, prompt builder, scorer, greedy inference settings, phase samples, checkpoint revisions, and raw generations for the standard question-answering sweeps.
The primary sweep contains 9{,}600 generations per system with no reported generation failure or scored null.
The synthetic check lacks raw generations and an environment manifest, and the matched comparison cannot be reconstructed from a recorded clean producing code state.

\runin{Limitations}
The study is limited to a small set of models, architectures, tasks, and prompts, and all ten-document results are exploratory.
Family comparisons retain differences in depth, positional encoding, state size, and checkpoint construction; the tighter hybrid contrast is uncertain.
The internal evidence is correlational, and the held-out split is unexamined.

\runin{Broader impact}
Position-sensitive evidence use can make retrieval-augmented systems unreliable and privilege facts because of placement rather than relevance.
Long-context accuracy should not be interpreted as uniform evidence access.

\runin{Conclusion and future directions}
Position bias varies across the evaluated sequence-mixer families and settings rather than following one universal long-context curve.
The shared end advantage and family-dependent beginning behavior also show why a single aggregate long-context score is incomplete.
Future work should train matched Transformer, state-space, and hybrid systems under one data and positional setup, then test targeted interventions on how evidence is retained and used.
Preregistered evaluation on the held-out questions is needed to confirm the present exploratory results.
Extending paired position curves to longer contexts, retrieval-augmented generation, and multi-step tasks would test whether these patterns predict downstream reliability.

\clearpage
\appendix

\section{Detailed experimental evidence}

The appendix preserves the analyses omitted from the four-page paper.
The main claims and their limitations do not depend on appendix-only evidence.

\subsection{Phase and artifact ledger}

\begin{table}[H]
\centering
\caption{Executed evidence by phase.
Counts are questions unless otherwise stated.}
\label{tab:phase-ledger}
\scriptsize
\setlength{\tabcolsep}{4.5pt}
\begin{tabular}{@{}clll@{}}
\toprule
Phase & Executed comparison & Committed source & Interpretation \\
\midrule
1 & 200 QA; 50 key-value & \texttt{artifacts/phase1} & Calibration and control \\
2 & 3 systems, $n=800$ each & \texttt{artifacts/phase2} & Primary family interaction \\
3 & pure and hybrid 8B, $n=800$ & \texttt{artifacts/phase3} & Matched, uncertain contrast \\
4 & 5 size pairs, $n=800$ each & \texttt{artifacts/phase4} & Capability and scale trend \\
5 & 2 corpora, $n=800$ each & \texttt{artifacts/phase5} & Corpus scope check \\
6 & 3 systems, 50 at two lengths & \texttt{artifacts/phase6} & Synthetic task check \\
7 & 100--200 per study; 1,200 states & \texttt{artifacts/phase7-mechanisms} & Hypothesis generation \\
8 & 3 systems, $n=800$ each & \texttt{artifacts/phase8} & Descriptive prevalence \\
\bottomrule
\end{tabular}
\end{table}

The dataset snapshot has 2{,}655 questions: 800 assigned to exploration and 1{,}855 assigned to confirmation.
The confirmatory questions have not been evaluated.

\subsection{Calibration and positive control}

\begin{figure}[H]
\centering
\includegraphics[width=0.92\linewidth]{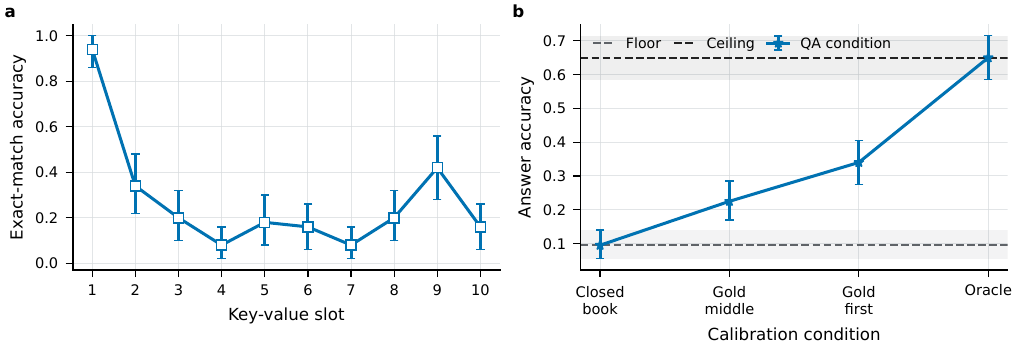}
\caption{Calibration of the primary Transformer setting.
The key-value task confirms that the pipeline recovers an imposed position signal, and the QA anchors separate guessing, the multi-document conditions, and answerability.}
\label{fig:app-calibration}
\end{figure}

The no-gold floor is $0.095$, the gold-only ceiling is $0.650$, and moving the gold document from the middle to the first position changes accuracy by $+0.115$ with 95\% CI $[0.055,0.175]$.
The key-value control produces a $+0.38$ edge-versus-middle effect.

\subsection{Primary family comparison}

The shared-corpus comparison uses one Transformer and two state-space checkpoints at similar parameter counts.
The Transformer primacy edge is $+0.052$ with 95\% CI $[0.032,0.072]$.
The two state-space primacy edges are $-0.001$ with CI $[-0.016,0.014]$ and $-0.018$ with CI $[-0.034,-0.003]$.
Their recency edges are $+0.075$, $+0.077$, and $+0.105$, respectively.
The paired state-space-minus-Transformer primacy interactions are $-0.053$ and $-0.070$, both Holm-corrected $p<10^{-4}$.
The models share training corpus and approximate scale, but not depth, positional encoding, state size, or checkpoint family.

\subsection{Matched hybrid comparison}

The pure model's primacy edge is $+0.008$ with CI $[-0.009,0.024]$, and the hybrid's is $+0.027$ with CI $[0.008,0.047]$.
The paired hybrid-minus-pure primacy effect is $+0.019$ with CI $[-0.006,0.044]$ and Holm $p=0.144$.
The paired recency effect is $-0.028$ with CI $[-0.055,-0.001]$ and Holm $p=0.089$.
The paired interaction is the relevant architecture test; different within-model significance levels do not establish a between-model difference.

\subsection{Capability and scale}

\begin{figure}[H]
\centering
\includegraphics[width=0.92\linewidth]{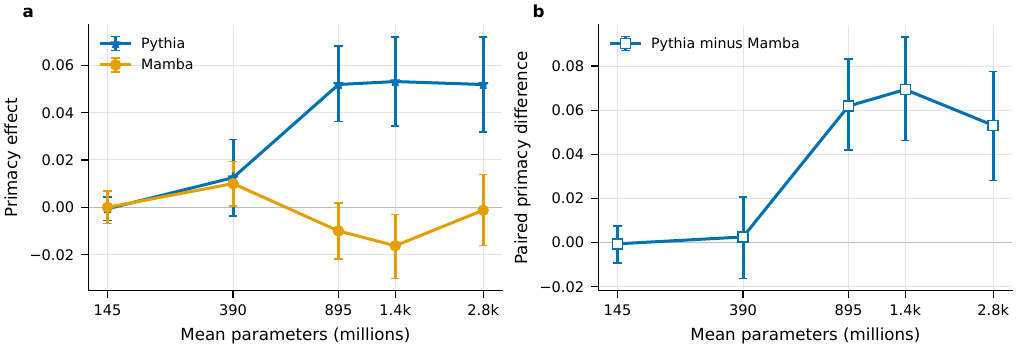}
\caption{Primacy across five approximate Transformer-state-space size pairs.
The family gap is visible only among models that can answer the task, so the trend combines capability and family.}
\label{fig:app-scale}
\end{figure}

Across the five pairs, the Transformer-minus-state-space primacy gap is near zero at the two smallest scales and is $+0.062$, $+0.069$, and $+0.053$ in the three larger pairs.
The smallest Transformer ceiling is only $0.009$, which prevents interpretation as a depth threshold or causal scale transition.

\subsection{Corpus and task checks}

\begin{figure}[H]
\centering
\includegraphics[width=0.62\linewidth]{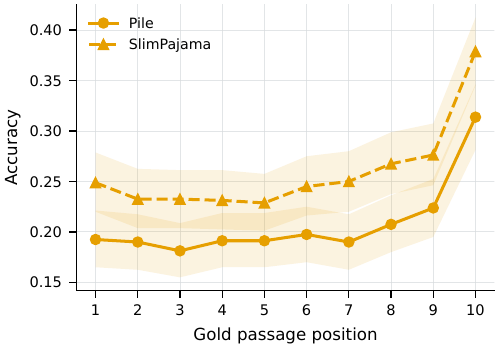}
\caption{Fixed architecture with two training corpora.
The corpus swap changes overall accuracy without a detectable edge-shape change.}
\label{fig:app-corpus}
\end{figure}

\begin{figure}[H]
\centering
\includegraphics[width=0.96\linewidth]{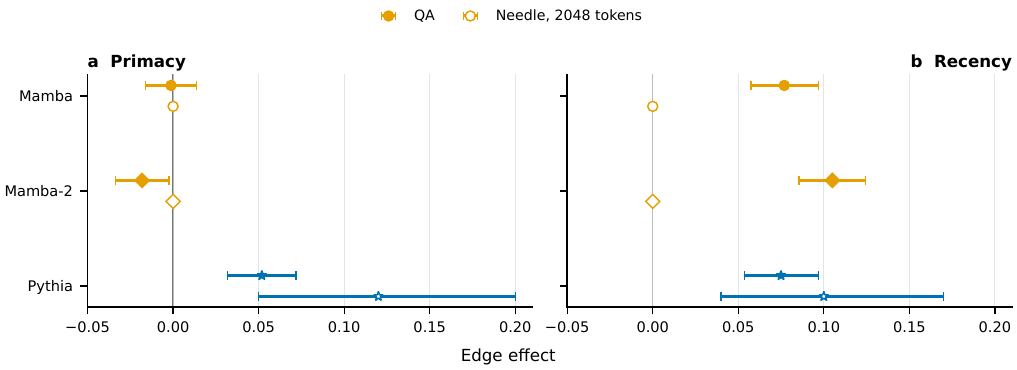}
\caption{Primacy and recency on multi-document QA and synthetic retrieval.
The synthetic task reproduces Transformer primacy but leaves the state-space systems at ceiling.}
\label{fig:app-task}
\end{figure}

Holding the state-space architecture fixed, the Pile-minus-SlimPajama primacy effect is $-0.007$ with CI $[-0.030,0.016]$ and the recency effect is $-0.016$ with CI $[-0.044,0.012]$.
On 50 RULER single-needle examples at 2K tokens, Transformer primacy is $+0.12$ with CI $[0.05,0.20]$, while both state-space models score $1.0$ at every tested depth.
Raw generations and an environment manifest for this synthetic check are not available in the repository snapshot.

\subsection{Prompt and internal measurements}

On 200 exploratory questions, repeating the question before and after the documents changes state-space primacy from $0.000$ to $+0.103$.
Placing the question only before the documents yields primacy of $+0.020$ and reduces recency from $+0.070$ to $+0.010$.
An alternative template reduces Transformer primacy from $+0.070$ to $+0.005$.
A balanced linear probe predicts edge versus middle position above a shuffled-label control in both families, but this does not establish answer-content storage or causal use.

\subsection{Additional systems}

\begin{figure}[H]
\centering
\includegraphics[width=0.60\linewidth]{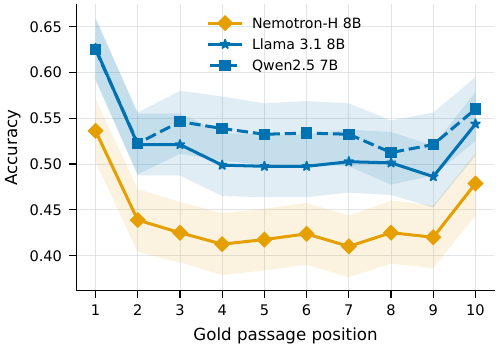}
\caption{Descriptive position curves for three additional Transformer and hybrid systems.
All show positive primacy, but cross-system differences do not isolate architecture.}
\label{fig:app-systems}
\end{figure}

The additional systems have primacy edges of $+0.076$, $+0.041$, and $+0.067$.
They differ in architecture, training data, tokenizer, token count, depth, and positional encoding, so the comparison establishes prevalence rather than mechanism.

\section{Controls and reproducibility}

The released key-value positive control detects a $+0.38$ edge-versus-middle effect.
On the 200-question exploratory Transformer sample, sham-gold primacy is $+0.005$ with CI $[-0.0325,0.0425]$, and distractor-order permutation accuracy has a maximum spread of $0.0233$.
These controls certify the paired position harness only in that setting and do not cover the matched 8B checkpoints.
The blinded audit has no human labels, and some control outputs are absent outside the primary setting.

The repository records the dataset split, prompts, scorer, greedy inference settings, phase samples, model revisions, raw generations for standard QA sweeps, and paired-bootstrap analysis.
The evaluation used the project compute nodes documented with the released artifacts and required at least approximately 50 single-GPU hours, excluding failed runs, reruns, plotting, and development.

\begin{table}[H]
\centering
\caption{Execution environments recorded by the phase artifacts.}
\label{tab:compute}
\scriptsize
\begin{tabular}{@{}lll@{}}
\toprule
Phases & Single-GPU node & Scope \\
\midrule
1 and 5 & A10G (23 GB) & Calibration and corpus check \\
2--4 & A40 (46 GB) & Primary, matched, and scale comparisons \\
6 & T4 (15 GB) and L40S (46 GB) & Synthetic retrieval \\
7 & L40S (46 GB) and T4 (15 GB) & Prompt, sink, and probe studies \\
8 & L40S (46 GB) & Additional systems \\
\bottomrule
\end{tabular}
\end{table}

\subsection{Checkpoint provenance}

\begin{table}[H]
\centering
\caption{Selected checkpoint repositories and pinned revisions.}
\label{tab:checkpoints}
\scriptsize
\begin{tabular}{@{}p{0.24\linewidth}p{0.45\linewidth}p{0.20\linewidth}@{}}
\toprule
Paper key & Repository identifier & Revision \\
\midrule
Transformer 2.8B & \texttt{EleutherAI/pythia-2.8b} & \texttt{2a259cdd96a4} \\
State-space v1 2.8B & \texttt{state-spaces/mamba-2.8b-hf} & \texttt{96c48e0292b6} \\
State-space v2 2.7B & \texttt{AntonV/mamba2-2.7b-hf} & \texttt{ef542707386f} \\
Pure state-space 8B & \texttt{nvidia/mamba2-8b-3t-4k} & \texttt{b915550c63ba} \\
Hybrid state-space 8B & \texttt{nvidia/mamba2-hybrid-8b-3t-4k} & \texttt{35e8852e2240} \\
SlimPajama state-space 2.8B & \texttt{state-spaces/mamba-2.8b-slimpj} & \texttt{a7bdd41af90c} \\
Nemotron-H 8B & \texttt{nvidia/Nemotron-H-8B-Base-8K} & \texttt{94ea861e008c} \\
Llama 3.1 8B & \texttt{meta-llama/Llama-3.1-8B} & \texttt{d04e592bb4f6} \\
Qwen2.5 7B & \texttt{Qwen/Qwen2.5-7B} & \texttt{d14972939875} \\
\bottomrule
\end{tabular}
\end{table}

\end{document}